\documentclass[a4paper]{article}

\usepackage{INTERSPEECH2019}

\title{Essence Knowledge Distillation for Speech Recognition}
\name{Zhenchuan Yang$^{1,\#}$, Chun Zhang$^{1,\#}$  \thanks{\# Both authors contribute equally to this work.}, Weibin Zhang$^{1,*}$, Jianxiu Jin$^1$, Dongpeng Chen$^2$}
\address{
  $^1$South China University of Technology, GuangZhou, China\\
$^2$VoiceAI Technologies Co. Ltd., Shenzhen, China}
\email{eeweibin@scut.edu.cn}

\begin{document}

\maketitle
\begin{abstract}
It is well known that a speech recognition system that combines multiple acoustic models trained on the same data significantly outperforms a single-model system. Unfortunately, real time speech recognition using a whole ensemble of models is too computationally expensive. In this paper, we propose to distill the knowledge of essence in an ensemble of models (i.e. the teacher model) to a single model (i.e. the student model) that needs much less computation to deploy. Previously, all the soften outputs of the teacher model are used to optimize the student model. We argue that not all the outputs of the ensemble are necessary to be distilled. Some of the outputs may even contain noisy information that is useless or even harmful to the training of the student model. In addition, we propose to train the student model with a multitask learning approach by utilizing both the soften outputs of the teacher model and the correct hard labels. The proposed method achieves some surprising results on the Switchboard data set. When the student model is trained together with the correct labels and the essence knowledge from the teacher model, it not only significantly outperforms another single model with the same architecture that is trained only with the correct labels, but also consistently outperforms the teacher model that is used to generate the soft labels.

\end{abstract}

\section{Introduction}
Automatic speech recognition (ASR), especially near-field speech recognition, has achieved great progress in recent years \cite{Mohamed2011Acoustic,Dahl2011Context}. But the problem of low-resource (i.e. limited training data) speech recognition is ubiquitous since a large amount of annotated data is not available for most languages used in the world. How to train an accurate model with limited training data remains a challenging problem. Various methods have been proposed to fully utilize the limited training data to improve the recognition accuracy of the speech recognition system trained on it.

Data augmentation has been shown to be an simple yet effective approach to increase the quantity and diversity of the data \cite{ragni2014data}. It has almost become part of the standard pipeline in data pre-processing for speech recognition. Corrupting clean data with noise \cite{kinoshita2013reverb}, vocal tract length perturbation (VTLP \cite{jaitly2013vocal}), speed-perturbation \cite{ko2015audio} have been widely adopted to improve the performance.

Model fusion is another technique to combine information at the other level of the acoustic modeling pipeline. By combining neural networks with distinct, complementary architectures \cite{deng2014ensemble,lu2014ensemble,fukuda2017efficient}, the whole system is able to capitalize on each architecture’s strengths to improve the system accuracy. However, usually it is impossible to deploy such a fused model to a large number of users since the computation is too expensive, especially if the individual models are large and complicated neural nets such as Long Short-Term Memory (LSTM \cite{Sak2014Long}) and Convolutional LSTM Deep Neural Networks (CLDNN \cite{Sainath2015Convolutional}). An effective method to deal with this problem is Knowledge Distillation (KD \cite{Hinton2015Distilling}).

Knowledge Distillation (KD) is a special transfer learning technology. It involves training a new model which is usually called the $ \it{student} $ or $ \it{student\ model} $. The knowledge in a well-trained model (the $ \it{teacher\ model} $, usually an ensemble of models) is compressed and transferred to the student model. This student model can be trained on a separate transfer data set without correct labels or on the original training data set where correct labels are available.

Various strategies have been proposed to compress and distill the knowledge of the teacher model to the student model. In the simplest form, the knowledge from the teacher model is distilled to the student model by training it with soften output probabilities produced by the teacher model \cite{Hinton2015Distilling,chebotar2016distilling}. Instead of combining the outputs of individual teacher model from an ensemble, Fukuda et.al. \cite{fukuda2017efficient} proposed to update the parameters of the student by randomly switching teacher labels at the minibatch level, or to train the student model on multiple output labels from various teacher models.
Huang et.al. \cite{huang2018knowledge} proposed to use sequence-level knowledge distillation instead of frame-level knowledge distillation. Similar sequence-level distillation was also explored in \cite{kanda2018sequence}. In the above research, none of the distilled student models are able to outperform their corresponding teacher models. There are still a big gap between the teacher models and the student models in terms of recognition accuracy. In our opinion, one of the reasons is that only the outputs of the teacher model are used to train the student model. The other reason might be that the student model is driven to learn everything from the teacher model, including not only valuable information that is beneficial to the generalization of the student model but also garbage information that is noisy and harmful.

To deal with this problem, we propose to train the student model with a multitask learning approach in order to utilize the correct hard labels. In addition, we propose to only distill the essence knowledge of the teacher model to the student model. One obvious problem is how to select the knowledge produced by the teacher model? We propose a simple yet effective method to select valuable information from the outputs of the teacher model. The student model is trained with correct labels and the selected knowledge from the teacher model. Surprisingly, the student model (with much less model parameters than the teacher model) trained with the proposed method consistently outperforms the teacher model.

The rest of this paper is organized as follows: Section 2 introduces the technologies used in our experiments, including data augmentation, model fusion and knowledge distillation. We introduce a method to select salient information for the outputs of the teacher model. The experimental setup and results are presented in Section 3. Finally, a simple conclusion is drawn in Section 4.

\section{Methodology}
The first step in knowledge distillation is to find a good teacher model. Then the knowledge of the teacher model is distilled into the student. In this section, we will firstly introduce data augmentation and model fusion that we use to train our teacher model. Then we will elaborate knowledge distillation, especially essence knowledge distillation.
\subsection{Data augmentation and model fusion}
Various strategies have been proposed for data augmentation in the training of deep neural networks, such as corrupting clean speech with noise or reverberations, vocal tract length perturbation (VTLP \cite{jaitly2013vocal}) and speed perturbation. According to the experiments in \cite{ko2015audio}, speed-perturbation technique gives more improvements than VTLP on various LVCSR tasks. Thus speed-perturbation is used for data augmentation in this paper. Speed-perturbation produces a time-warped speech signal, $x(at)$, from the original signal $x(t)$. $a$ is a factor that modifies the speed of the speech signal. When $a>1$, the warped speech will be shifted in spectrogram towards high frequency, and the duration of the resulting speech will be reduced. On the other hand, when $a<1$, $x(at)$ will sound slower than the original speech $x(t)$. Due to the change in the length of the signal, the alignments for the speed perturbed data need to be regenerated. In this paper, we expand the original training data 3 times by setting $a$ to 0.9, 1.0 and 1.1.

Model fusion is an effective method to combine individual models into an ensemble of models, which usually performs much better by merging those models' predictions \cite{fukuda2017efficient}. To obtain an ensemble of models, we need to separately train several sub-models.
One of the most important questions about model fusion is how to choose the sub-models to make the fused model work best? Ideally, we want the sub-models to compensate one another so that the fused model has a lower error rate than any individual model \cite{Kuncheva2003Measures, Banfield2005Ensemble}. In fact, as demonstrated in \cite{deng2014ensemble, Soltau2014Joint, Sainath2013Improvements}, it is better to fuse sub-models with different network architectures in order to increase the diversity of the fused model and make less error rate. In our experiment, sub-models of different architectures are used to form the teacher model.

Given $K$ sub-models that are trained with different architectures, one way to get the outputs of the fused model is to use the weighted average of the frame-level posteriors produced by all the sub-models. In our preliminary experiments, we found that another equally good method is to combine the inputs to the $ softmax $ layers of sub-models by taking their weighted average, and then feed the combined input to a $ softmax $ layer to produce the final predictions. That is, for the logits (the inputs to the final $ softmax $) of the $i$-th output node of the $k$-th sub-model, $z_{ik}$, the outputs of the fused model is computed as:
\begin{equation}
\begin{aligned}
\begin{split}
\label{eq:fusion}
q_i &= softmax(z_i/T)\\
z_{i}&=  \sum_{k=1}^K w_kz_{ik}
\end{split}
\end{aligned}
\end{equation}
where $w_k \in [0,1] $, $\sum_{k=1}^K w_k=1$ are the weighting parameters. The best weighting parameters $w_k$ for model fusion can be found through grid searching. $T$ is a temperature parameter to adjust the “softness” of the output distribution. In a simple classification task (e.g. a handwritten digits recognition task) where probabilities produced by the teacher model are all very small (e.g. $10^{-7}$) except the one for the correct answer. We can increase the temperature $T$ to produce a softer probability (e.g $10^{-7}$ might be increased to 0.01) distribution over classes, but after training, the temperature is set to be 1.

For the sub-models to be able to be fused together using the above method, their outputs should have the same dimensionality. In speech recognition, this can be achieved by constructing the individual models with the same decision tree.

\subsection{Knowledge Distillation}
As mentioned above, Knowledge Distillation has achieved great success in many aspects. It involves training a new model (i.e. the student model) by using the output probabilities produced by the teacher model as “soft targets” for the student model. Since the teacher is able to generate a soft target distribution given any input sample, the training of the student model can be carried out on a data set without any label. However, in this paper, we only consider the case where the correct labels are available since we will train the student model on the original training data. In this case, we propose to formulate the knowledge distillation process in a multitask learning framework in order to utilize the original correct labels.

The first learning task for the student model is to learn to produce the correct labels. Let $ \theta $ denotes the parameters of the student model to be updated. The objective function for this task is the cross entropy with the correct labels and can be formulated as below:
\begin{equation}
\label{cross-entropy}
J_{CE}(\theta)=-\sum_{i=1}^Cy_i\log v_i
\end{equation}
where C is the size of ground-truth targets, $ y_i $ is the hard label and $v_i $ is the output probability of the class of the student model. In practice, usually one-hot label is used, so Equation (\ref{cross-entropy}) is  equivalent to:
\begin{equation}
J_{CE}(\theta)=-\log v_c
\end{equation}
where $v_c$ is the output probability generated by the student model for the correct label.

In the second learning task, the student model is trained to emit posteriors that are as close as possible to those of the teacher model. To achieve this, the Kullback-Leibier (KL) divergence between the posterior distributions generated by the teacher and student models is minimized. In other words, we need to minimize:
\begin{equation}
\begin{aligned}
D_{KL}(q_i||v_i) &= \sum_{i}q_i\log\frac{q_i}{v_i} \\
&=\sum_{i}-q_i\log v_i + \sum_{i}q_i\log q_i \\
&= H(q,v)-H(v)\\
\end{aligned}
\end{equation}
where
\begin{equation}
\begin{aligned}
\begin{split}
H(q,v)&=\sum_{i}-q_i\log v_i \\
H(v)&=\sum_{i}q_i\log q_i
\end{split}
\end{aligned}
\end{equation}
and $q_i$ are the soft labels generated by the teacher model given the same input. $H(q,v)$ is the cross entropy between the outputs of the student model and the soft labels. $H(v)$ is independent of the parameters of the student model and thus can be omitted from the above equation. Therefore, the second objective function for the student model is:
\begin{equation}
J_{KD}(\theta)=-\sum_{i=1}^Cq_i\log v_i
\end{equation}
Combining the two learning tasks, the overall objective function for the student model can be formulated as:
\begin{equation}
\label{final-obj}
J_{L}(\theta)=\lambda J_{CE}(\theta)+(1-\lambda)J_{KD}(\theta)
\end{equation}
where $ \lambda \in [0,1] $ is a  parameter to adjust the importance of the two parts. When $\lambda=0$, the student model is trained purely based on the soft labels generated by the teacher model. This is the approach used in \cite{chebotar2016distilling,fukuda2017efficient}. When $\lambda=1$, no knowledge is distilled into the student model.

\subsection{Essence knowledge distillation}
Compared with the traditional method where the student is trained only with the correct labels, the soft labels generated by the teacher model provide valuable information that defines a rich similarity structure over the data. For example, in a handwritten digit recognition task, given one version of a 3 as input for training, Table 1 shows the one-hot hard label and a possible soften output generated by the teacher model. Except the correct answer indicated by the hard label (i.e. the correct label), no further information is available since the probabilities for other answers are all zero. However, from the soften labels produced by the teacher model, we can see that the number 3 looks more like 2 and 8 than other numbers. The similarity structure may be learned by the student model to help generalize better on separated testing data sets.
\begin{table}[t]
  \caption{The one-hot hard label and a possible soften output generated by the teacher model given one version of a 3 as input for training in a handwritten digit recognition task.}
  \renewcommand\tabcolsep{1.5 pt}
  \centering
  \begin{tabular}{l|c|c|c|c|c|c|c|c|c|c c}
    \hline
    digit & 0 & 1 & 2 & 3 & 4 & 5 & 6 & 7 & 8 & 9   \\
    \hline
    Hard-label  &  0& 0& 0& 1& 0& 0& 0& 0& 0& 0&       \\
    \hline
    Soft-label  &  0.02& 0.02& 0.1& 0.7& 0.03& 0.01& 0.01& 0.01& 0.08& 0.02&   \\
    \hline
  \end{tabular}
\end{table}

\begin{figure}[t]
  \centering
  \includegraphics[width=\linewidth]{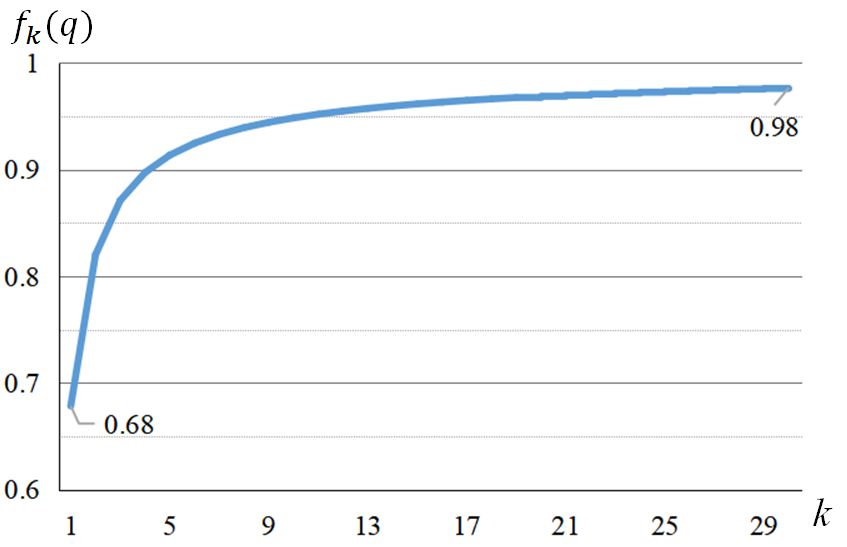}
  \caption{The average summation of the top $k$ largest output probabilities ($f_k(q)$) with different number of $k$ computed in a sentence in speech recognition.}
  \label{fig:top-k}
\end{figure}

Mathematically, the $softmax$ layer will always generate probabilities that are larger than zero. Unfortunately, not all these non-zero numbers are valuable. In the above example, the output probability of the number 4 is larger than the output probability of the number 6, but this does not mean that the number 4 looks more like the number 3 than the number 6. In speech recognition where the dimensionality of the output layer is several thousand or even tens of thousands, much of the valuable information resides in a small number of the output nodes. In our experiments, the dimensionality of the output layer is 8912. Let $f_k(q)$ denotes the summation of the top $k$ largest values of the output $q$. Figure \ref{fig:top-k} shows the average value of $f_k(q)$ with different number of $k$ computed in a sentence with the temperature $T$ set to 1. As can be seen, the average probability for the top-1 answer is about 0.68. The summarized probability for the top-10 answer is about 0.9. When $k$ goes to 40, the summarized probability is about 0.98, meaning that the summarized probability for the remaining outputs (i.e. about 8912-40=8872 nodes) is only about 0.02.

\begin{figure}[t]
  \centering
  \includegraphics[width=\linewidth]{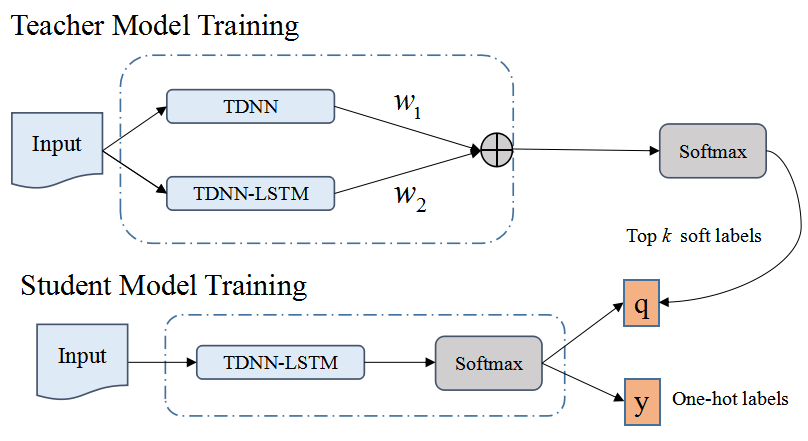}
  \caption{The whole process of the proposed training method. The teacher model is an ensemble of models that are trained on the same data set. The logits from each individual model are fused and then fed into a softmax layer. The top $k$ largest probabilities (i.e. top $k$ soft labels) generated by the teacher model, together with the correct labels, are used to train the student model in a multitask learning framework. }
  \label{fig:kd}
\end{figure}

We argue that these non-zero little probabilities generated by the teacher model even with a high temperature $T$ are noisy, useless and even harmful to the training of the student model. Therefore, we proposed to only keep the top $k$ outputs of the teacher model and set others to zero. That’s what we call the essence of knowledge from the output of the teacher model. The resulting vector will then be renormalized to make it a probability distribution. Finally the renormalized vector will serve as the soft labels for knowledge distillation in Equation (\ref{final-obj}). Another advantage of using the top $k$ outputs is that the training of the student model will be much faster. The whole process of the proposed training method is shown in Figure \ref{fig:kd}.

\section{Experiment}
The proposed essence knowledge distillation was evaluated on the 309-hour Switchboard English conversational telephone speech task. Word error rates (WER) are presented on the Hub5’00 evaluation set that contains 20 conversations from Switchboard (SWBD) and 20 conversations from CallHome English (CHE). We used speed perturbation technique described in \cite{cui2015data} to augment the data 3-fold. A 30k-vocabulary 4-gram language model trained from the transcription of the Switchboard corpus and interpolated with the Fisher corpus was used for decoding.

The Kaldi \cite{povey2011kaldi} toolkit was used to conduct all the experiments. 40-dimensional Mel-frequency cepstral coefficients (MFCC) extracted every 10ms from the input speech were used as inputs to the neural networks. 100 dimensional i-vectors were also provided to perform speaker adaptation of the networks. The i-vector also provides information about the mean of the speaker’s data and thus no cepstral truncation is needed. Stochastic gradient descent (SGD) with an exponentially decreasing learning rate schedule was used to update all the model parameters. The cross entropy was used as the objective function. The model averaging technique proposed in \cite{Povey2014Parallel} was used for parallel training.

Time delay neural networks (TDNN) and hybrid TDNN-LSTM (long short term memory) neural networks were used in our experiments. All the neural networks were constructed using the same decision tree to ensure that they all have 8912 output nodes for model fusion and knowledge distillation. In addition, all the TDNN neural networks (including those in a TDNN-LSTM hybrid neural network) were configured to have a left context of -16 and a right context of 12.

As to using model fusion to construct the teacher model, it is well known that increasing sub-model’s diversity help improve the fused model’s performance. Initially we trained different sub-models with different speech speed (i.e. with different $a$ for time-warped speech signal $x(at)$) and then fused these sub-models together to form the teacher model. But this did not work very well. We then focused on increasing acoustic model diversity, especially acoustic models with different neural network architectures. In the following experiments, a TDNN model and a TDNN-LSTM model were fused as the teacher model for knowledge distillation. The fusing weight (i.e. the $w_k$ in Equation (\ref{eq:fusion})) is set to 0.5.

As shown in Figure \ref{fig:top-k}, usually the teacher model does not output very high probability for the top-1 answer in speech recognition. The average probability for the top-1 answer is about 0.68. We found in our preliminary experiments that there was no need to increase the temperature $T$ of the teacher model for knowledge distillation. The output of the teacher model is soft enough to be directly used for distillation. Therefore, in the following knowledge distillation experiments, the temperature $T$ was always set to 1.

\begin{table}[t]
  \caption{Word error rates of different models trained with a subset of the Switchboard data. }
  \label{tab2}
  \renewcommand\tabcolsep{3.0pt}
  \centering
  \begin{tabular}{lccccc}
    \toprule
    Acoustic Model    & \emph{k}  & SWB   & CHE & TOTAL  \\
    \midrule
    TDNN                 &   NA    &14.1  &26.3   &20.3   \\
    TDNN-LSTM            &   NA &14.4  &26.2   &20.2   \\
    \midrule
    TDNN-LSTM+TDNN (teacher)    &  NA    &13.2  &25.4   &19.3   \\
    \midrule
                    &  1&       13.5  &25.4  &19.6   \\
                    &  5&       13.1  &24.6  &18.9   \\
                    & 10&      13.0  &24.6  &\textbf{18.8}   \\
    TDNN-LSTM (student)   &  20&       12.9 &25.0  &19.0   \\
                     &  50&       13.0 &24.9  &18.9   \\
                   & 1000&      13.0& 24.8 &19.0    \\
                   & 8912&      13.0 & 24.6 & 18.9    \\
    \bottomrule
  \end{tabular}
\end{table}

A subset consisting 25\% of the training data from the Switchboard data set was used to quickly evaluate the effectiveness of the proposed method and to tune some hyperparameters. The results are shown in Table \ref{tab2}. As can be seen, the TDNN-LSTM performed better than the TDNN model. This result coincides with previous findings \cite{cheng2017exploration}. The teacher model, which is a fusion of a TDNN model and a TDNN-LSTM model, significantly outperformed any individual model.

For knowledge distillation, when we substantially increased the number of output nodes to be distilled to the student model (e.g. $k=1000$), the performance of the student model may even get worse. Therefore, to perform essence knowledge distillation, it is desirable to use $k$ with relatively small values. We can see that when $k$ is set with a relatively small value, the student model can even outperform the corresponding teacher model given the correct labels that are also available. In addition, when $k$ is small, the training of the student model is much faster. When we are writing this manuscript, the training of the student model when $k=8912$ (i.e. all the output nodes from the teacher model are distilled to the student model) is not finished.

We then went on to perform experiments on the whole Switchboard data set and the results are shown in Table \ref{tab3}. The same conclusion with that on the Switchboard subset can be drawn.

\begin{table}[t]
  \caption{Word error rates of different models trained with the whole Switchboard data set.}
  \label{tab3}
  \renewcommand\tabcolsep{3.0pt}
  \centering
  \begin{tabular}{lccccc}
    \toprule
    Acoustic Model    & \emph{k}  & SWB   & CHE & TOTAL  \\
    \midrule
    TDNN        &    NA&12.4    &24.0    &18.3       \\
    TDNN-LSTM   &     NA&12.1     &23.7 &17.9        \\
    \midrule
    TDNN-LSTM+TDNN (teacher)  &   NA& 11.6 & 22.7 &17.2    \\
    \midrule
      &   1&       11.6 & 22.5 &17.1   \\
     &   5&       11.2&  22.0 &\textbf{16.7}   \\
     TDNN-LSTM (student)&   10&      11.3&  22.2 &16.8   \\
      &   20&      11.4&  22.1 &16.8   \\
      &   50&      11.4&  22.1 &16.8  \\
    \bottomrule
  \end{tabular}
\end{table}

\section{Conclusions and future work}
In this paper, we propose to distill the essence knowledge from a teacher model to a student model. The outputs of the $softmax$ layer in a neural network will always be positive. However, not all the non-zero soft labels produces by the teacher model are valuable to be distilled to the student model. We propose to only select the top $k$ values (i.e. the essence knowledge) from the teacher model. Our experiments on the Switchboard data show that student models trained with essence knowledge distillation can surprisingly outperform the teacher model.

In the future, it would be very interesting to explore the idea of iteratively training better student models and better teacher models to see when and how the method converges. We would also like to apply the same idea in discriminative training.

\bibliographystyle{IEEEtran}

\bibliography{mybib}

\end{document}